\documentclass[11pt]{article}

\usepackage[final]{acl}

\usepackage{times}
\usepackage{latexsym}

\usepackage[T1]{fontenc}

\usepackage[utf8]{inputenc}

\usepackage{microtype}

\usepackage{inconsolata}

\usepackage{graphicx}

\usepackage{inconsolata}

\usepackage{graphics}
\usepackage{xspace,mfirstuc,tabulary,}
\usepackage{amssymb, amsmath}
\usepackage{array,booktabs, multirow}
\usepackage{CJKutf8}
\usepackage{tabularx}
\usepackage{dsfont} %

\newcolumntype{Y}{>{\centering\arraybackslash}X}

\newcommand{\model}{$\mathcal{M}$\xspace}

\newcommand{\porig}{$p(s,r)$\xspace}

\newcommand{\ptokens}{\texttt{P-Tokens}\xspace}

\newcommand{\bedit}{\texttt{BEGIN\_EDIT}\xspace}
\newcommand{\eedit}{\texttt{END\_EDIT}\xspace}

\newcolumntype {+}{ >{\global\let\currentrowstyle\relax}}
\newcolumntype {^}{ >{\currentrowstyle }}

\newenvironment{courier}{%
    \fontsize{7}{7}\fontfamily{pcr}\selectfont 
}{%
    \par 
}

\title{Persuasion Tokens for Editing Factual Knowledge in LLMs}

\author{Paul Youssef \hspace{0.25cm} Christin Seifert  \hspace{0.25cm} Jörg Schlötterer \\ 
Marburg University 
\\ \texttt{\{paul.youssef, christin.seifert, joerg.schloetterer\}@uni-marburg.de}}

\begin{document}
\maketitle
\begin{abstract}
In-context knowledge editing (IKE) is a promising technique for updating Large Language Models (LLMs) with new information. However, IKE relies on lengthy, fact-specific demonstrations which are costly to create and consume significant context window space. In this paper, we introduce persuasion tokens (\ptokens) -- special tokens trained to replicate the effect of IKE demonstrations, enabling efficient knowledge editing without requiring fact-specific demonstrations. We evaluate \ptokens across two editing datasets and three LLMs, demonstrating performance comparable to, and often exceeding, IKE. We further find that editing performance is robust to distractors with small negative effects to neighboring facts, and that increasing the number of \ptokens improves performance. Our work addresses key limitations of IKE and provides a more practical and scalable alternative for editing LLMs.\footnote{\url{https://github.com/paulyoussef/p-tokens}}

\end{abstract}

\section{Introduction}
Large Language Models (LLMs) encode facts about the world in their parameters~\cite{petroni-etal-2019-language, youssef-etal-2023-give, youssef2024enhancing}. 
Knowledge editing methods (KEs) have been introduced to address the problem of outdated factual knowledge in LLMs~\cite{wang-etal-2024:ACMSurvey, mazzia2024surveyknowledgeeditingneural}. KEs differ in how they update knowledge in LLMs. Parameter-modifying KEs~\cite{meng2023massediting, tan2023massive} directly change the model's parameters, whereas parameter-preserving KEs~\cite{wang2024wise, guo2025balancedit} introduce additional memory modules to update knowledge without affecting the original parameters. 

In-context knowledge editing (IKE)~\cite{zheng-etal-2023-edit} makes use of the strong in-context learning abilities of LLMs and updates knowledge in LLMs by adding the updated knowledge as input to the model. More specifically, IKE constructs a prompt that contains the new fact with a set of 32 demonstrations to teach the model to use the new fact in semantically similar contexts, and to not affect other irrelevant facts. IKE has been shown to have strong performance. The diverse demonstrations, which are the reason for the strong performance of IKE, bring also key limitations. These demonstrations are fact-specific and have to be constructed from scratch for each edit. These demonstrations also make the prompt longer, which consumes much of the context window and makes inference slower. 

\begin{figure}
     \centering
     \includegraphics[width=1\columnwidth, trim={0cm 7cm 0cm 0cm}]{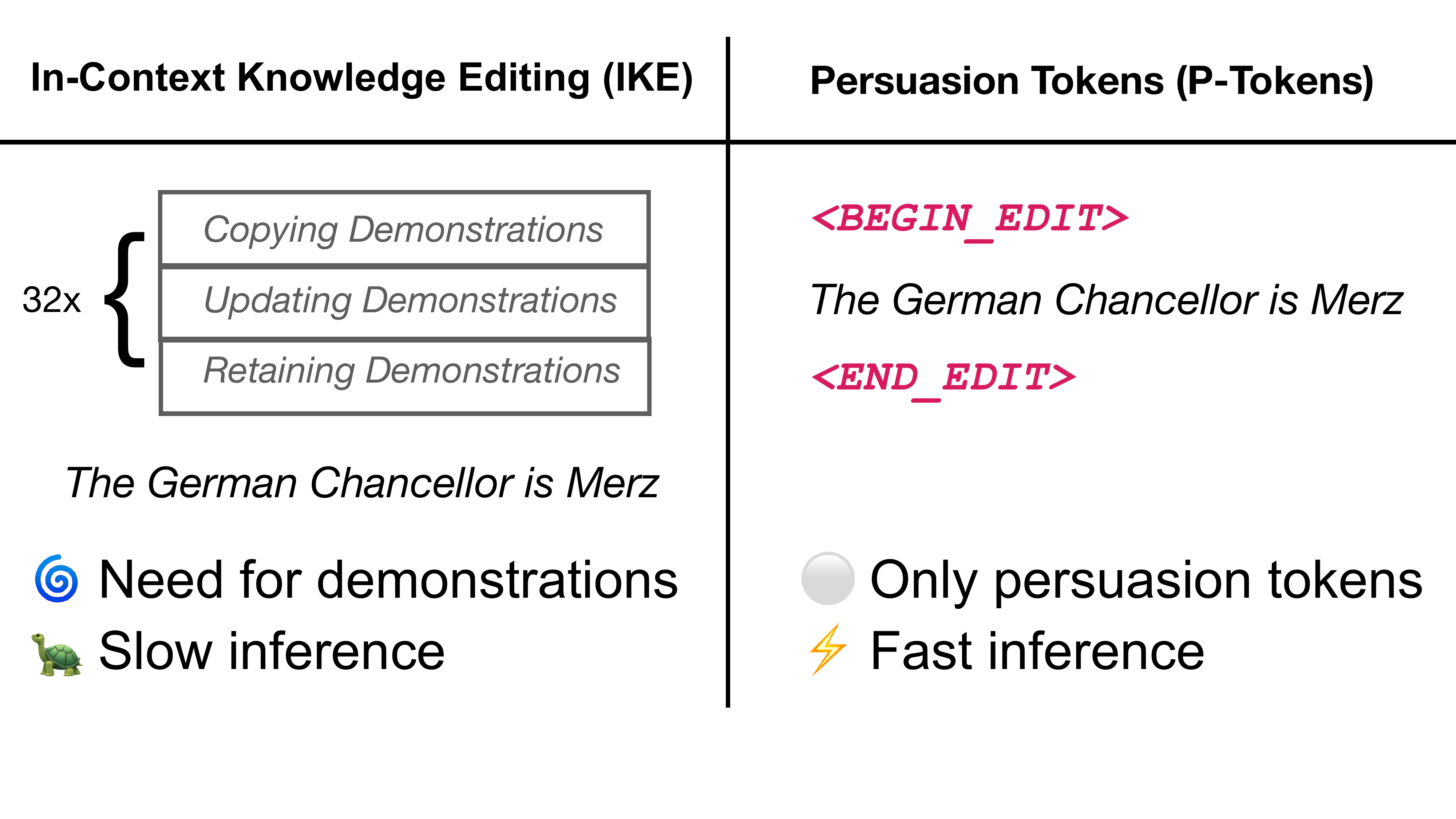}
     \caption{In-context knowledge editing (IKE) relies on complex demonstrations and leads to slower inference. Persuasion tokens (\ptokens) eliminate the need for long demonstrations and lead to faster inference.}
     \label{fig:fig1}
 \end{figure}

We introduce persuasion tokens (\ptokens), which are special tokens that are tuned to have the same effect as the IKE demonstrations and can edit facts in-context.
\ptokens eliminate the need for fact-specific demonstrations and replace the long IKE edits, leading to faster inference (see Figure~\ref{fig:fig1}). We train and evaluate \ptokens on two standard knowledge editing datasets and three LLMs. Our experiments show strong performance of \ptokens compared to IKE across all settings. Performance improves, with increasing the number of \ptokens. We further study the effect of distractors (additional edits) on the editing performance. We find that the performance on editing and paraphrased prompts remains robust to distractors, but neighboring facts are negatively affected. We further compare the number of tokens and inference time for IKE and \ptokens. Compared to IKE, \ptokens require less tokens and have less inference time, highlighting the efficiency of \ptokens. Our work provides an efficient, effective and a more practical alternative to IKE, paving the way for wider adoption. %

\section{Background and Problem Statement}

\paragraph{In-context Knowledge Editing.}
Facts are often represented as triplets of $(subject, relation, object)$, or $(s, r, o)$ for short. Querying an LLM with a prompt $p(s,r)$, where $p$ expresses the relation $r$ with the subject $s$ (e.g., ``In which city is the Eiffel Tower?''), should lead to generating the object $o$ (e.g., ``Paris''), given that the fact $(s, r, o)$ is encoded in the LLM. A knowledge editing operation $E(s, r, o, o', p)$ is successful if it changes the output of the LLM such that the retrieved object is $o^{\prime}$ instead of $o$. IKE edits are conducted by prepending an editing prompt $p_{{}_{IKE}}(s,r,o')$ to the query prompt $p(s,r)$, i.e., $p_{{}_{IKE}} (s,r,o') \oplus p(s,r)$, where $\oplus$ is the string concatenation operation. The editing prompt $p_{{}_{IKE}}(s,r,o')$ causes the LLM's output to change from $o$ to $o'$.

\paragraph{Problem Statement.}
Given a model \model that outputs the object $o$, when provided with a prompt \porig, i.e., $o = argmax_{o*} \ \mathbb{P}_{\mathcal{M}(p(s,r))} [o^*]$, where $\mathbb{P}_{\mathcal{M}(p(s,r))}$ is the model's output probability distribution over the vocabulary $\mathcal{V}$ given the prompt $p(s,r)$ and $o^* \in \mathcal{V}$. \model's output is changed to $o'$ with IKE~\cite{zheng-etal-2023-edit} by prepending an IKE editing prompt $p_{{}_{IKE}}(s,r,o')$ (see Figure~\ref{fig:ike_example} in the appendix for an example) to the original prompt, i.e., $o' = argmax_{o^*} \ \mathbb{P}_{\mathcal{M}(p_{{}_{IKE}}(s,r,o')\oplus p(s,r))} [o^*]$. Our goal is to replace the editing prompt $p_{{}_{IKE}}(s,r,o')$ with a significantly shorter prompt $p_{{}_{PT}}(s,r,o')$ such that $o' = argmax_{o^*} \ \mathbb{P}_{\mathcal{M}(p_{{}_{PT}}(s,r,o')\oplus p(s,r))} [o^*]$. Accordingly, the shorter prompt $p_{{}_{PT}}(s,r,o')$ should have the same effect as the IKE editing prompt $p_{{}_{IKE}}(s,r,o')$, and lead to generating $o'$.

\section{Method}
\label{sec:method}
To replace the IKE prompt with a short and efficient editing prompt, we enclose the edit $p(s,r,o')$ with \emph{special editing tokens:} \texttt{BEGIN\_EDIT} and \texttt{END\_EDIT}. We refer to these tokens as \emph{persuasion tokens} or \ptokens for short. We further optimize the embedding vectors of these tokens to minimize the Kullback-Leibler (KL) divergence loss between two output distributions: 

\begin{equation}
	\mathcal{L} = KL[P_{_{PT}}  || \  P_{_{IKE}}] 
\end{equation}

where $P_{_{PT}} = \mathbb{P}_{\mathcal{M}(p_{{}_{PT}}(s,r,o')\oplus p(s,r))}$ is the output distribution of the model when using the above described editing prompt with \ptokens, and $P_{_{IKE}} = \mathbb{P}_{\mathcal{M}(p_{{}_{IKE}}(s,r,o')\oplus p(s,r))}$ is the output distribution when using the long IKE editing prompt. To improve performance, we further minimize the KL-divergence between the output distributions that correspond to the following pairs of inputs:

\paragraph{Paraphrases:} $p_{{}_{PT}}(s,r,o')\oplus p'(s,r)$ and $p_{{}_{IKE}}(s,r,o')\oplus p'(s,r)$, where $p'(s,r)$ refers to a paraphrased version of the prompt $p(s,r)$.

\paragraph{Neighbors:} $p_{{}_{PT}}(s,r,o')\oplus p(\bar{s},r)$ and $p(\bar{s},r)$. Here, we use $p(\bar{s},r)$ to refer to neighboring prompts, that query similar facts to the one targeted by the edit (same relation, but different subject). Our goal is to avoid affecting irrelevant facts that should not be changed by the edit.  
    
\paragraph{Distractors:} We add a distractor that consists of several edits (with \ptokens) between the edit $p_{{}_{PT}}(s,r,o')$, and the querying prompt $p(s,r)$, to enhance robustness.
    
\paragraph{Other:} We add \ptokens to other prompts that do not include any edits, such that \ptokens have no effect when there are no edits. 

We provide an overview of all of the pairs of inputs in Tables~\ref{tab:opt_pairs:cf} and ~\ref{tab:opt_pairs:zsre} in Appendix~\ref{app:eval}. 
Each training batch includes all types of input pairs.

\section{Experimental Setup}

\paragraph{Datasets.} We use two datasets in our experiments. The first is CounterFact~\cite{meng2022locating}, which contains counterfactual statements (e.g., ``The space needle is located in Rome.''). We use the same subset of 2,000 examples as \citet{zheng-etal-2023-edit}. We use 800 of these examples for training, 200 for validation and the remaining 1,000 for testing. The second dataset is zsRE~\cite{levy-etal-2017-zero, mitchell2022fast}, which contains question-answer pairs. From the training set, we randomly sample 800 instances for training and 200 for validation. For testing, we use the whole test set of 19,086 examples. There are no IKE edits for this dataset. As a replacement for IKE edits, we use a baseline containing the editing prompt or the paraphrased prompt. An overview of the examples used for zsRE is shown in Table~\ref{tab:opt_pairs:zsre} in Appendix~\ref{app:eval}.

\paragraph{Evaluation.} We follow previous work~\cite{meng2022locating, meng2023massediting} in using dataset-specific metrics. For CounterFact, the evaluation is based on the probability of the original object $o$ and the edited object $o'$. An edit is considered successful if the probability of $o'$ is higher than the probability of $o$. 
The metric \textbf{ES} refers to the editing success rate when the editing prompts and the prompts to retrieve the edited fact are the same, whereas \textbf{PS} refers to the success rate when the prompts to retrieve the edited fact are paraphrased. A corresponding metric for neighboring facts is \textbf{NS}, where the goal is to not affect these prompt by retaining a higher probability for $o$ than $o'$. The overall performance is summarized by \textbf{S}, which is the harmonic mean of \textbf{ES}, \textbf{PS} and \textbf{NS}. For zsRE, the metrics are based on the accuracy of the model in generating $o'$ using the editing prompts (\textbf{Efficacy}), the paraphrased prompts (\textbf{Paraphrase}), and the accuracy of the model on unrelated facts (\textbf{Specificity}). We provide more formal definitions in Appendix~\ref{app:eval}.

\paragraph{Baselines.} We compare against IKE~\cite{zheng-etal-2023-edit} to asses if \ptokens are as effective. IKE demonstrations are only available for CounterFact. Therefore, we cannot report IKE performance on zsRE. We instead compare against a baseline that contains the edit $p(s,r,o')$ without \ptokens, and also report results on the original model without any edits. 

\paragraph{Number of \ptokens.} To analyze how the number of tokens affects performance, we experiment with $m$ persuasions tokens,  $m \in \{1, 3, 5, 7, 10\}$. This means, we use $m$ \bedit tokens and $m$ \eedit tokens. Each of the $2m$ tokens has its own separate representation.

\paragraph{Models.} We use GPT-J-6B~\cite{gpt-j}, Qwen2.5-7B, Qwen2.5-14B~\cite{qwen2.5} and Llama3-8B~\cite{grattafiori2024llama3herdmodels}. 

\section{Results and Discussion}

\begin{figure*}[t]
\centering
\includegraphics[width=0.9\columnwidth]{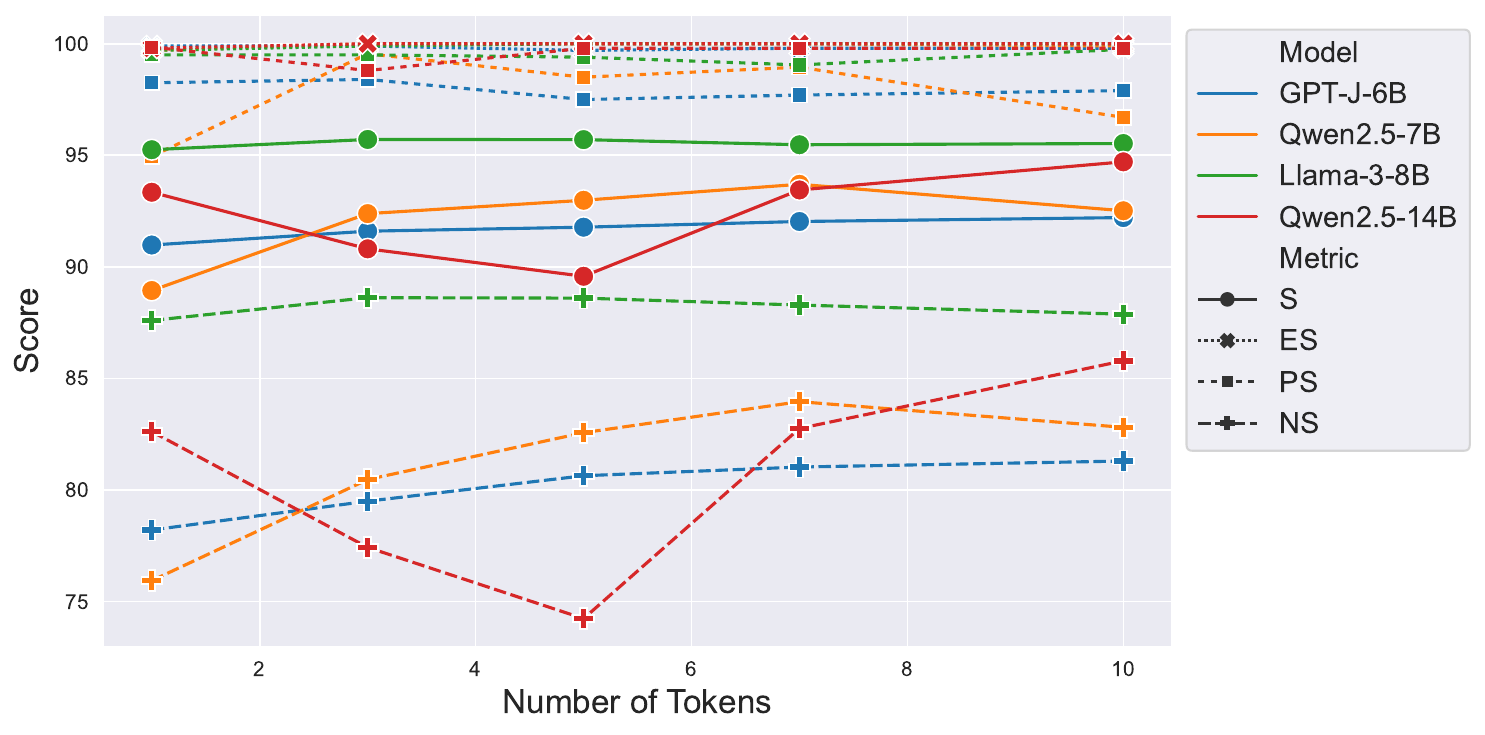}
\includegraphics[width=\columnwidth]{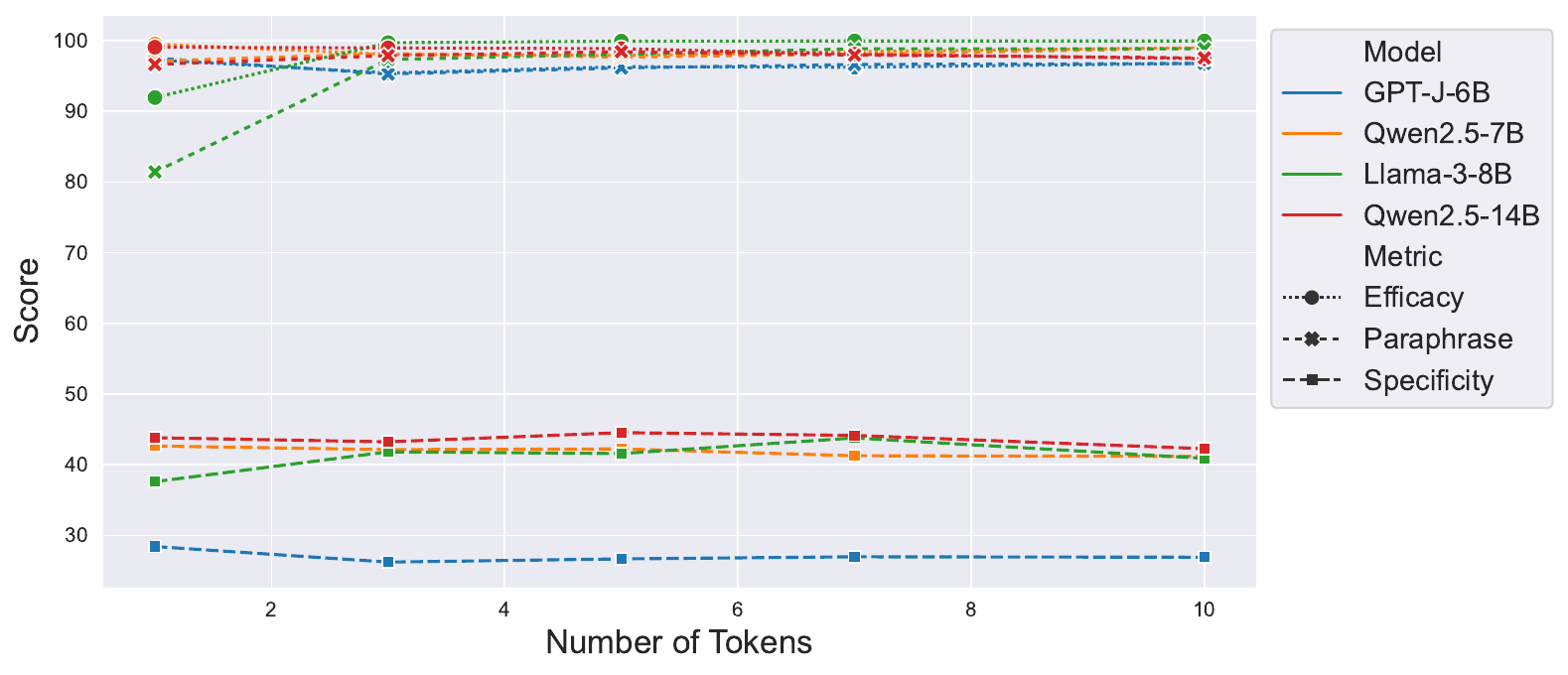}

\caption{Performance across different numbers of P-Tokens (Left: CounterFact, Right: zsRE). On CounterFact, Qwen2.5-7B benefits from increasing the number of tokens, especially on \textbf{PS} and \textbf{NS}. On zsRE, Llama3's \textbf{Efficacy} and \textbf{Paraphrase} increase when the number of P-Tokens increases. }%
\label{fig:num_tokens}
\end{figure*}

The results on CounterFact are shown in the left part of Table~\ref{tab:overall}. We report the maximum performance across different numbers of \ptokens. More detailed results can be found in Appendix~\ref{app:results}. \ptokens outperform IKE across all models. For example, the \textbf{S} value on Llama3 with IKE is $93.97$, whereas with \ptokens it is $95.82$. On Qwen-7B, the gap is even larger ($\approx 5$ p.p.). Generally, the performance of \ptokens is on par or slightly better than the performance of IKE with respect to \textbf{ES} and \textbf{PS}. This shows that \ptokens are as effective as IKE in editing knowledge. %
We notice a larger gap with respect to \textbf{NS} ($\approx 4$ p.p. on Llama3, Qwen-14B and GPT-J, and $\approx 10$ p.p. on Qwen-7B), which indicates that \ptokens have minimum effects on neighboring facts. We attribute this to our optimization criteria (cf. Table~\ref{tab:opt_pairs:cf} in Appendix A), where we optimize the outputs with \ptokens on unrelated facts to be the same as the outputs of the original model, which also explains the small gap to the original model ($1.58$ p.p. at most).

\begin{table*}[]
    \centering
\resizebox{0.9\textwidth}{!}{%
\begin{tabular}{cl@{\hskip 5mm}cccc@{\hskip 10mm}ccc}

\toprule
 & & \multicolumn{4}{c}{\textbf{CounterFact}} & \multicolumn{3}{c}{\textbf{zsRE}} \\ \cmidrule(r{2.5em}){3-6}\cmidrule{7-9}

          \textbf{Model} &     \textbf{KE} &     \textbf{S}&     \textbf{ES} &     \textbf{PS} &    \textbf{NS} &  \textbf{Efficacy} &  \textbf{Paraphrase} &  \textbf{Specificity} \\
\midrule

    \multirow{4}{*}{\rotatebox[origin=c]{90}{GPT-J}} & original & 24.49 &  17.10 & 19.25 & \textbf{82.70} &     27.23 &       26.43 &        27.22 \\
     & baseline & 74.02 &  95.90 & 87.90 & 53.41 & 93.36 &       87.96 &        25.39 \\
     & IKE & 89.96 & \textbf{100} & 97.30 & 76.51 &  \textemdash & \textemdash  &  \textemdash \\
     & \ptokens & \textbf{92.39} &  99.90 & \textbf{98.40} & 81.30 & \textbf{97.29} &       \textbf{97.54} &        \textbf{28.44} \\ \midrule

     \multirow{4}{*}{\rotatebox[origin=c]{90}{Qwen-7B}} &     original & 21.45 &  13.80 & 17.95 & \textbf{85.54} &   39.50 &       37.91 &        39.51 \\
      & baseline & 62.07 &  99.80 & 90.40 & 36.70 & 96.41 &       89.16 &        \textbf{44.18} \\
      &      IKE & 89.08 & \textbf{100} & 98.75 & 73.79 & \textemdash & \textemdash  &  \textemdash \\
      & \ptokens & \textbf{93.88} & \textbf{100} & \textbf{99.55} & 83.96 &     \textbf{99.51} &       \textbf{98.94} &        42.66 \\ \midrule
      
\multirow{4}{*}{\rotatebox[origin=c]{90}{Llama3}}  &     original & 12.14 &   7.50 &  9.75 & \textbf{89.36} &     38.46 &       37.19 &        39.15 \\
 & baseline & 72.42 & \textbf{100} & 89.35 & 49.43 &     98.73 &       95.14 &        43.49 \\
 &      IKE & 93.97 & \textbf{100} & 99.25 & 84.40 & \textemdash & \textemdash  &  \textemdash \\
 & \ptokens & \textbf{95.82} & \textbf{100} & \textbf{99.75} & 88.62 &     \textbf{99.96} &       \textbf{98.85} &        \textbf{43.80} \\ \midrule

\multirow{4}{*}{\rotatebox[origin=c]{90}{Qwen-14B}}  
&  original & 17.27 &  10.60 & 14.70 & \textbf{87.82} &     42.34 &       40.74 &        42.16 \\
& baseline & 63.86 &  99.90 & 90.65 & 38.56 &     94.01 &       87.51 &        \textbf{45.04} \\
&      IKE & 92.84 & \textbf{100} & 99.30 & 81.68 & \textemdash & \textemdash  &  \textemdash \\
& \ptokens & \textbf{94.72} & \textbf{100} & \textbf{99.85} & 85.79 &     \textbf{99.07} &       \textbf{98.44} &        44.55 \\
\bottomrule
\end{tabular}}
\caption{Performance on CounterFact and zsRE. \ptokens outperform IKE and the baseline across all models.}
\label{tab:overall}
\end{table*}

\begin{table}[t]
    \centering
\resizebox{\columnwidth}{!}{%
\begin{tabular}{cccccc}
\toprule
          \textbf{Model} &  $ \lvert \textbf{Dist.} \rvert $&     \textbf{S} &    \textbf{ES} &    \textbf{PS} &    \textbf{NS} \\
\midrule
       \multirow{4}{*}{\rotatebox[origin=c]{90}{GPT-J}} 
               &      0 & 92.79 &  99.80 & 98.05 & 82.57 \\
                &     10 & 92.86 &  99.90 & 98.55 & 82.31 \\
                &     50 & 92.18 &  99.80 & 98.90 & 80.55 \\
                &    100 & 92.37 &  99.90 & 98.85 & 80.95 \\ \midrule
     \multirow{4}{*}{\rotatebox[origin=c]{90}{Qwen-7B}} 
                &      0 & 94.10 & 100 & 99.45 & 84.55 \\
                &     10 & 92.76 & 100 & 99.65 & 81.27 \\
                &     50 & 90.85 & 100 & 99.55 & 77.06 \\
                &    100 & 89.86 & 100 & 99.40 & 75.04 \\ \midrule
\multirow{4}{*}{\rotatebox[origin=c]{90}{Llama3}} 
                &      0 & 96.36 & 100 & 99.55 & 90.18 \\
                &     10 & 95.59 & 100 & 99.80 & 88.00 \\
                &     50 & 93.59 & 100 & 99.30 & 83.44 \\
                &    100 & 89.55 & 100 & 98.80 & 74.75 \\ \midrule 

\multirow{4}{*}{\rotatebox[origin=c]{90}{Qwen-14B}} 
                 &      0 & 94.88 & 100.0 & 99.90 & 86.15 \\
                 &     10 & 93.72 & 100.0 & 99.90 & 83.33 \\
                 &     50 & 91.81 & 100.0 & 99.80 & 79.01 \\
                 &    100 & 88.87 & 100.0 & 99.90 & 72.75 \\

\bottomrule
\end{tabular}}
    \caption{Performance of \ptokens on a subset of CounterFact when adding distractors of varying length. $\lvert \textbf{Dist.} \rvert$ refers to the distractor length. \textbf{ES} and \textbf{PS} remain high, while \textbf{NS} drops, indicating that neighbouring facts are affected more as the distractor becomes longer. }
    \label{tab:distractor}
\end{table}

The right part of Table~\ref{tab:overall} shows the results for zsRE. We observe that \ptokens outperform the baseline in most cases. The gaps in performance differ across metrics. With \textbf{Efficacy}, the differences vary between $\approx 1$ p.p. (Llama3) and $\approx 4$ p.p. (Qwen and GPT-J). With \textbf{Paraphrase} the differences are larger and vary between $\approx 4$ p.p. (Llama3) and $\approx 10$ p.p. (Qwen and GPT-J). The high \textbf{Paraphrase} values are due to our optimization criteria (cf. Table~\ref{tab:opt_pairs:zsre} in Appendix A), where we target approximating a model's output, whose input includes the edit with a paraphrased prompt. With \textbf{Specificity} the values differ slightly from the values of the original models, indicating minor effects to neighboring facts.   

\paragraph{Number of \ptokens.} We show how the performance varies across different numbers of \ptokens for CounterFact and zsRE in Figure~\ref{fig:num_tokens}. On CounterFact, we notice that Qwen-7B benefits the most from increasing the number of tokens. Qwen-7B's \textbf{PS} and \textbf{NS} increase by $\approx 5$ p.p. when using $3$ instead of $1$ tokens. On zsRE, Llama3's \textbf{Efficacy} and \textbf{Paraphrase} increase the most by $\approx 7$ and $16$ p.p. respectively, when increasing the number of tokens from $1$ to $3$. Generally, the results show that editing performance improves when increasing the number of tokens. %

\paragraph{Effect of distractors.} We evaluate the performance of \ptokens with distractors of varying length on a subset of 1,000 facts from CounterFact. 
The results in Table~\ref{tab:distractor} show that \textbf{ES} and \textbf{PS} scores remain high, indicating successful editing on both the editing and paraphrased prompts.
The overall performance (\textbf{S} score) drops by $\approx 5$ and $7$ p.p. on Qwen and Llama3 respectively, while it remains relatively stable on GPT-J. 
The \textbf{ES} and \textbf{PS} scores remain high indicating successful editing on both the editing and paraphrased prompts. 
However, the \textbf{NS} scores (and accordingly the overall scores \textbf{S}) drop significantly on Qwen-7B (10 p.p.), Qwen-14B (14 p.p.) and Llama3 (15 p.p.). This drop suggests that distractors mostly affect neighboring facts. A potential remedy might be increasing the distractor length used in training (we use distractors consisting of 5 or 10 edits). 

\paragraph{Ablation.} We retrain \ptokens for Llama3 without distractors to verify their effect on the performance. The results in Table~\ref{tab:distractor:ablation} show that the performance of \ptokens drops when training without distractors. For example, the \textbf{S} score drops from 89.55 (cf. Table~\ref{tab:distractor}) to 82.94 when having a distractor of 100 edits, which demonstrates the effectiveness of having distractors during training for a more robust performance. 

\begin{table}[t]
    \centering
\resizebox{\columnwidth}{!}{%
\begin{tabular}{cccccc}
\toprule
          \textbf{Model} &  $ \lvert \textbf{Dist.} \rvert $&     \textbf{S} &    \textbf{ES} &    \textbf{PS} &    \textbf{NS} \\
\midrule
\multirow{4}{*}{\rotatebox[origin=c]{90}{Llama3}} 
 &      0 & 96.33 & 100.0 & 99.35 & 90.28 \\
 &     10 & 94.75 &  99.9 & 95.80 & 89.18 \\
 &     50 & 87.93 &  98.4 & 84.20 & 82.78 \\
 &    100 & 82.94 &  91.9 & 80.25 & 77.95 \\

\bottomrule
\end{tabular}}
    \caption{Performance of \ptokens, trained \emph{without} distractors, on a subset of CounterFact when adding distractors of varying length. The performance of \ptokens drops compared to our proposed setup (cf. Table~\ref{tab:distractor}).  }
    \label{tab:distractor:ablation}
\end{table}

\begin{table}[!h]
\small
    \centering
    \resizebox{\columnwidth}{!}{%
        \begin{tabular}{lcc}
        \toprule
        \textbf{Method} & \textbf{\#Tokens} & \textbf{Inference Time per Edit}\\
        \midrule
        IKE&959.19&0.17\\

        \ptokens&58.01&0.03\\
        \bottomrule
        \end{tabular}}
    \caption{Comparison of IKE and \ptokens with respect to the number of tokens and inference time in seconds, using Qwen2.5-7B with $10\times2$ \ptokens and a sample of 1000 edits. Inference with \ptokens is over five times faster than with IKE.}
    \label{tab:token_time_comparison}
\end{table}

\paragraph{Efficiency.} Part of the motivation for \ptokens is to make inference more efficient. We compare the number of tokens and the inference time for IKE and \ptokens using Qwen-7B  with $10\times2$ \ptokens, a sample of 1,000 edits and a batch size of 1 for both methods. The results in Table~\ref{tab:token_time_comparison} show that IKE prompts are more than 16 times as long as \ptokens prompts. Inference with \ptokens is over five times faster than with IKE. 
The shorter prompts of \ptokens also enable larger batch sizes, increasing the inference time advantage even more.

\paragraph{Amortization point.} Despite the advantages at inference time, \ptokens add costs for the initial training. We calculate at which point the training costs amortize with Qwen-7B. Training Qwen-7B with $10\times2$ \ptokens takes roughly 15 hours and 28 minutes. This means that \ptokens amortize after $(15 \times 60 \times 60 + 28\times60)/(0.17 - 0.03)=398\text{k}$ inferences. 
If the expected number of inferences is lower, training \ptokens is not economical, but still has other advantages like eliminating the need for edit-specific demonstrations and requiring shorter prompts.

\section{Related Work}
Knowledge Editing methods (KEs)~\cite{wang-etal-2024:ACMSurvey, mazzia2024surveyknowledgeeditingneural} can be classified as either parameter-modifying or parameter-preserving. Parameter-modifying KEs directly change the model's parameters to update facts, and can further be categorized as locate-and-edit KEs~\cite{meng2022locating,meng2023massediting}, that first locate the parameters responsible for the facts and then adapt these parameters, or meta-learning KEs~\cite{mitchell2022fast, tan2023massive} that train hypernetworks to predict the necessary shift in parameters to update facts. Parameter-preserving KEs add special memory modules that update the targeted facts~\cite{mitchell2022memory, hartvigsen2023aging, wang2024wise, guo2025balancedit}, or make use of the model's in-context abilities~\cite{cohen-etal-2024-evaluating, youssef-etal-2024-queen} to update some facts using some demonstrations~\cite{zheng-etal-2023-edit}. Our work eliminates the need for fact-specific demonstrations and makes editing easier and more efficient using \ptokens. %

\section{Conclusion}
In this work, we introduced persuasion tokens as an alternative to in-context knowledge edits that require many demonstrations and overload the model's context. We showed that persuasion tokens outperform in-context edits on two datasets and three LLMs. We further showed that increasing the number of persuasion tokens has a positive effect on the performance, and that distractors negatively affect the performance on neighboring facts, while the performance on editing and paraphrased prompts remain robust to distractors. \ptokens are an efficient and effective alternative to IKE, and their integration in LLMs enables easy editing. %

\section*{Limitations}
Persuasion tokens can be used as an alternative to IKE demonstrations and lead to faster inference. However, these tokens need to be trained initially, which incurs additional computational costs. Knowledge Editing is shown to have potential malicious use cases~\cite{chen2024can, 10.1145/3711896.3737001, cheng2025finetuningeraseeditsfragile, youssef2025positioneditinglargelanguage, youssef-etal-2025-fact, youssef2025tracingreversingrankonemodel}. \ptokens could make prompt injection attacks easier due to the prompts being shorter. However, some work already addresses this risk~\cite{youssef-etal-2025-make, chen2025defending}.

\section*{Acknowledgments}
We gratefully acknowledge support from the hessian.AI Service Center (funded by the Federal Ministry of Research, Technology and Space, BMFTR, grant no. 16IS22091) and the hessian.AI Innovation Lab (funded by the Hessian Ministry for Digital Strategy and Innovation, grant no. S-DIW04/0013/003)

\bibliography{custom}

\appendix

\section{Evaluation and Implementation Details}
\label{app:eval}
\paragraph{Evaluation.} We follow previous work~\cite{meng2022locating, meng2023massediting} in  using dataset-specific metrics. 
For CounterFact, we use the following metrics: 
\begin{itemize}
    \item Efficacy Score (\textbf{ES}): the percentage of facts where the probability of the edited object is higher than the probability of the original object
    $ES = \frac{1}{n}\sum_i^{n} \mathds{1}[ \mathbb{P}_{\mathcal{M}(p_{{}_{PT}}(s,r,o')\oplus p(s,r))}(o') > \mathbb{P}_{\mathcal{M}(p_{{}_{PT}}(s,r,o')\oplus p(s,r))}(o)]$ where $\mathds{1}[\cdot]$ is the indicator function.
\item Paraphrase Score (\textbf{PS}): Similar to ES, but with paraphrased prompts, i.e., $PS = \frac{1}{n}\sum_i^{n} \mathds{1}[\mathbb{P}_{\mathcal{M}(p_{{}_{PT}}(s,r,o')\oplus p'(s,r))}(o') > \mathbb{P}_{\mathcal{M}(p_{{}_{PT}}(s,r,o')\oplus p'(s,r))}(o)]$. 

\item Neighborhood Score (\textbf{NS}): Proportion of neighboring (irrelevant) facts, for which the probability of the original object is still higher than that of the edited object, $NS = \frac{1}{n}\sum_i^{n} \mathds{1}[ \mathbb{P}_{\mathcal{M}(p_{{}_{PT}}(s,r,o')\oplus p(\bar{s},r))}(o') < \mathbb{P}_{\mathcal{M}(p_{{}_{PT}}(s,r,o')\oplus p(\bar{s},r))}(o)]$. 

\end{itemize}

For zsRE, we measure the accuracy of the model in generating the edited object using editing (\textbf{Efficacy}) and paraphrased prompts (\textbf{Paraphrase}), 
$\frac{1}{n}\sum_i^{n} \mathds{1}[ o' = argmax_{o^*}\mathbb{P}_{\mathcal{M}(p_{{}_{PT}}(s,r,o')\oplus p/p'(s,r))}(o*)]$, where $ p/p'(s,r)$ refers to using either $p(s,r)$ or $p'(s,r)$. We measure the accuracy on unrelated facts (\textbf{Specifcity})
$\frac{1}{n}\sum_i^{n} \mathds{1}[\bar{o} = argmax_{o^*}\mathbb{P}_{\mathcal{M}(p_{{}_{PT}}(s,r,o')\oplus p(\bar{s},\bar{r}))}(o*)]$.

\paragraph{Optimization pairs.} We show the optimization prompt pairs for CounterFact and zsRE in Table~\ref{tab:opt_pairs:cf} and Table~\ref{tab:opt_pairs:zsre} respectively.

\paragraph{Implementation details.} We optimize the embeddings of the \ptokens using AdamW~\cite{loshchilov2018decoupled} optimizer with $\beta_1 = 0.9$ and $\beta_2 = 0.98$ with weight decay=$0.01$. We train for a maximum of 50 epochs using early stopping with with a patience of 3 epochs on the validation set. We use a distractor consisting of 5 or 10 facts for each edit

\paragraph{Datasets.} CounterFact~\cite{meng2022locating} is published under the  MIT License. We could not find the license for zsRE~\cite{levy-etal-2017-zero}.

\begin{table*}
\centering
    \begin{tabular}{lcc}
    \toprule
\textbf{Name}&        \textbf{Persuasion Token Prompts} & \textbf{Target Prompts} \\ \midrule
    Editing prompts & $p_{{}_{PT}}(s,r,o')\oplus p(s,r)$ & $p_{{}_{IKE}}(s,r,o')\oplus p(s,r)$ \\

    Paraphrases &     $p_{{}_{PT}}(s,r,o')\oplus p'(s,r)$ & $p_{{}_{IKE}}(s,r,o')\oplus p'(s,r)$ \\

    Neighbors & $p_{{}_{PT}}(s,r,o')\oplus p(\bar{s},r)$ & $p(\bar{s},r)$ \\

    Distractor & $p_{{}_{PT}}(s,r,o') \oplus p_{{}_{PT}}(\bar{s}, \bar{r}, \bar{o}') \oplus p(s,r)$ & $p_{{}_{IKE}}(s,r,o')\oplus p(s,r)$ \\

    Only \bedit & $p_{{}_{PT}}(B)\oplus p(s,r)$ & $p(s,r)$ \\
    Only \eedit & $p_{{}_{PT}}(E)\oplus p(s,r)$ & $p(s,r)$ \\
    empty edit &  $p_{{}_{PT}}(B,E)\oplus p(s,r)$ & $p(s,r)$ \\
    empty edit reversed &  $p_{{}_{PT}}(E,B)\oplus p(s,r)$ & $p(s,r)$ \\
\bottomrule

    \end{tabular}
    \caption{Prompts used to train persuasion tokens with the CounterFact dataset. We optimize the embedding vectors of persuasion tokens to minimize the KL-divergence between the output distributions given the \textbf{Persuasion Token Prompts} and the output distributions given the \textbf{Target Prompts} (target distributions). }
    \label{tab:opt_pairs:cf}

\end{table*}

\begin{table*}
\centering
    \begin{tabular}{lcc}
    \toprule
\textbf{Name}&        \textbf{Persuasion Token Prompts} & \textbf{Target Prompts} \\ \midrule
    Editing prompts & $p_{{}_{PT}}(s,r,o')\oplus p(s,r)$ & $p(s,r,o') \oplus p(s,r)$ \\

    Paraphrases &     $p_{{}_{PT}}(s,r,o')\oplus p'(s,r)$ & $p'(s,r,o')\oplus p'(s,r)$ \\

    Neighbors & $p_{{}_{PT}}(s,r,o')\oplus p(\bar{s},\bar{r})$ & $p(\bar{s},\bar{r})$ \\
    Distractor & $p_{{}_{PT}}(s,r,o') \oplus p_{{}_{PT}}(\bar{s}, \bar{r}, \bar{o}') \oplus p(s,r)$ & $p(s,r,o') \oplus p(s,r)$ \\

    Only \bedit & $p_{{}_{PT}}(B)\oplus p(s,r)$ & $p(s,r)$ \\
    Only \eedit & $p_{{}_{PT}}(E)\oplus p(s,r)$ & $p(s,r)$ \\
    empty edit &  $p_{{}_{PT}}(B,E)\oplus p(s,r)$ & $p(s,r)$ \\
    empty edit reversed &  $p_{{}_{PT}}(E,B)\oplus p(s,r)$ & $p(s,r)$ \\
\bottomrule

    \end{tabular}
    \caption{Prompts used to train persuasion tokens with the zsRE dataset. We optimize the embedding vectors of persuasion tokens to minimize the KL-divergence between the output distributions given the \textbf{Persuasion Token Prompts} and the output distributions given the \textbf{Target Prompts} (target distributions). }
    \label{tab:opt_pairs:zsre}

\end{table*}

\section{Further Results}
\label{app:results}
We show the performance over different numbers of tokens for CounterFact and zsRE in Table~\ref{tab:num_tokens_overall}. We show the performance with different numbers of tokens and distractors in Table~\ref{tab:distractor_tokens}

\begin{table*}[]
    \centering
\resizebox{0.8\textwidth}{!}{%
\begin{tabular}{cc@{\hskip 5mm}cccc@{\hskip 10mm}ccc}

\toprule
 & & \multicolumn{4}{c}{\textbf{CounterFact}} & \multicolumn{3}{c}{\textbf{zsRE}} \\ \cmidrule(r{2.5em}){3-6}\cmidrule{7-9}

\textbf{Model} &    \textbf{\#Tokens} &     \textbf{S} &     \textbf{ES} &    \textbf{PS} &    \textbf{NS} &  \textbf{Efficacy} &  \textbf{Paraphrase} &  \textbf{Specificity} \\
\midrule
  \multirow{5}{*}{\rotatebox[origin=c]{90}{GPT-J}}  
    &        1 & 90.98 &  99.90 & 98.25 & 78.21 &  \textbf{97.29} &       \textbf{97.54} &        \textbf{28.44} \\
    &        3 & 91.60 &  99.90 & 98.40 & 79.50 &     95.43 &       95.28 &        26.24 \\
    &        5 & 91.78 &  99.70 & 97.50 & 80.64 &     96.31 &       96.14 &        26.69 \\
    &        7 & 92.03 &  99.80 & 97.70 & 81.03 &     96.24 &       96.63 &        26.99 \\
    &       10 & \textbf{92.21} &  \textbf{99.80 }& \textbf{97.90 }& \textbf{81.30} &     96.79 &       96.78 &        26.93 \\ \midrule
\multirow{5}{*}{\rotatebox[origin=c]{90}{Qwen-7B}}  
  &        1 & 88.94 &  99.70 & 94.95 & 75.94 &     \textbf{99.51} &       97.11 &        \textbf{42.66} \\

  &        3 & 92.39 & \textbf{100 }& 99.55 & 80.47 & 98.06 &       98.17 &        42.15 \\
  &        5 & 92.99 & \textbf{100 }& 98.50 & 82.57 & 97.89 &       97.67 &        42.25 \\
  &        7 & \textbf{93.70} & \textbf{100 }& \textbf{98.95 }& \textbf{83.96} &     98.45 &       98.10 &        41.28 \\
  &       10 & 92.52 &  99.90 & 96.70 & 82.81 &     99.01 &       \textbf{98.94} &        41.19 \\ \midrule
\multirow{5}{*}{\rotatebox[origin=c]{90}{Llama3}} 
  &        1 & 95.25 &  99.70 & 99.50 & 87.60 &     91.95 &       81.43 &        37.62 \\
  &        3 & 95.71 &  99.90 & 99.50 & \textbf{88.62} &     99.72 &       97.33 &        41.84 \\
  &        5 & \textbf{95.70} & \textbf{100 }& 99.40 & 88.60 &     99.93 &       98.07 &        41.59 \\
  &        7 & 95.47 & \textbf{100 }& 99.05 & 88.29 &     \textbf{99.96 }&       \textbf{98.85} &        \textbf{43.80} \\
  &       10 & 95.53 & \textbf{100 }& \textbf{99.75} & 87.88&     \textbf{99.96} &       \textbf{98.85} &        40.92 \\ \midrule
\multirow{5}{*}{\rotatebox[origin=c]{90}{Qwen-14B}} 
 &        1 & 93.34 &  99.80 & \textbf{99.85} & 82.61 &  \textbf{99.07} &       96.64 &        43.83 \\
 &        3 & 90.81 & \textbf{100} & 98.80 & 77.43 &  98.98 &       97.88 &        43.25 \\
 &        5 & 89.58 & \textbf{100} & 99.80 & 74.24 &  98.87 &       \textbf{98.44} &        \textbf{44.55} \\
 &        7 & 93.45 & \textbf{100} & 99.80 & 82.76 &  98.08 &       97.96 &        44.16 \\
 &       10 & \textbf{94.71} & \textbf{100} & 99.80 & \textbf{85.79} &  97.45 &       97.56 &        42.28 \\
  
\bottomrule
\end{tabular}}
    \caption{Editing performance on CounterFact with different number of persuasion tokens.}
    \label{tab:num_tokens_overall}
\end{table*}

\begin{table*}[]
    \centering
    \tiny
\begin{tabular}{lrrrrrr}
\toprule
          \textbf{Model} &   $ \lvert \textbf{Dist.} \rvert $ &  \textbf{\#Tokens} &     \textbf{S} &     \textbf{ES} &    \textbf{PS} &    \textbf{NS} \\
\midrule

       GPT-J &         0 &           1 & 91.41 &  99.80 & 98.05 & 79.37 \\
       GPT-J &         0 &           3 & 91.68 &  99.70 & 97.25 & 80.58 \\
       GPT-J &         0 &           5 & 92.00 &  99.40 & 97.10 & 81.64 \\
       GPT-J &         0 &           7 & 92.42 &  99.40 & 97.20 & 82.57 \\
       GPT-J &         0 &          10 & 92.04 &  99.00 & 96.90 & 82.15 \\
       GPT-J &        10 &           1 & 91.20 &  99.90 & 98.55 & 78.50 \\
       GPT-J &        10 &           3 & 92.05 &  99.90 & 97.80 & 80.94 \\
       GPT-J &        10 &           5 & 91.99 &  99.50 & 96.05 & 82.31 \\
       GPT-J &        10 &           7 & 92.27 &  99.80 & 97.30 & 81.87 \\
       GPT-J &        10 &          10 & 91.92 &  99.60 & 96.50 & 81.73 \\
       GPT-J &        50 &           1 & 88.56 &  99.80 & 98.90 & 72.76 \\
       GPT-J &        50 &           3 & 90.96 &  99.30 & 96.75 & 79.52 \\
       GPT-J &        50 &           5 & 83.79 &  88.00 & 83.15 & 80.55 \\
       GPT-J &        50 &           7 & 62.25 &  60.30 & 52.30 & 80.05 \\
       GPT-J &        50 &          10 & 27.60 &  20.20 & 21.45 & 79.49 \\
       GPT-J &       100 &           1 & 87.33 &  99.90 & 98.85 & 70.30 \\
       GPT-J &       100 &           3 & 24.92 &  17.50 & 19.65 & 80.95 \\
       GPT-J &       100 &           5 & 27.00 &  19.20 & 21.50 & 79.95 \\
       GPT-J &       100 &           7 & 27.66 &  20.50 & 21.20 & 79.88 \\  
       GPT-J &       100 &          10 & 26.92 &  19.40 & 21.15 & 79.25 \\ \midrule
     Qwen-7B &         0 &           1 & 88.86 &  99.80 & 94.20 & 76.20 \\
     Qwen-7B &         0 &           3 & 92.34 & 100 & 99.45 & 80.42 \\
     Qwen-7B &         0 &           5 & 93.39 & 100 & 98.55 & 83.50 \\
     Qwen-7B &         0 &           7 & 93.95 & 100 & 98.95 & 84.55 \\
     Qwen-7B &         0 &          10 & 92.57 & 100 & 96.50 & 83.01 \\
     Qwen-7B &        10 &           1 & 89.47 &  99.90 & 99.05 & 74.49 \\
     Qwen-7B &        10 &           3 & 89.97 & 100 & 99.65 & 75.13 \\
     Qwen-7B &        10 &           5 & 90.09 & 100 & 99.60 & 75.42 \\
     Qwen-7B &        10 &           7 & 92.30 &  99.80 & 98.25 & 81.27 \\
     Qwen-7B &        10 &          10 & 89.06 & 100 & 98.45 & 73.92 \\
     Qwen-7B &        50 &           1 & 88.23 & 100 & 96.85 & 73.12 \\
     Qwen-7B &        50 &           3 & 86.21 & 100 & 99.55 & 67.78 \\
     Qwen-7B &        50 &           5 & 90.18 & 100 & 99.40 & 75.73 \\
     Qwen-7B &        50 &           7 & 90.46 & 100 & 98.15 & 77.06 \\
     Qwen-7B &        50 &          10 & 82.39 & 100 & 98.90 & 61.35 \\
     Qwen-7B &       100 &           1 & 85.32 & 100 & 95.60 & 68.02 \\
     Qwen-7B &       100 &           3 & 83.94 & 100 & 99.40 & 63.78 \\
     Qwen-7B &       100 &           5 & 88.33 &  99.90 & 99.15 & 72.11 \\
     Qwen-7B &       100 &           7 & 89.43 &  99.90 & 97.95 & 75.04 \\
     Qwen-7B &       100 &          10 & 77.95 & 100 & 98.45 & 54.56 \\ \midrule
Llama3 &         0 &           1 & 95.77 &  99.80 & 99.45 & 88.89 \\
Llama3 &         0 &           3 & 96.25 &  99.80 & 99.40 & 90.18 \\
Llama3 &         0 &           5 & 96.07 &  99.90 & 98.95 & 90.01 \\
Llama3 &         0 &           7 & 95.70 &  99.80 & 98.55 & 89.45 \\
Llama3 &         0 &          10 & 96.14 & 100 & 99.55 & 89.60 \\
Llama3 &        10 &           1 & 95.06 & 100 & 99.50 & 86.90 \\
Llama3 &        10 &           3 & 95.00 &  99.80 & 99.45 & 86.94 \\
Llama3 &        10 &           5 & 94.75 &  99.90 & 99.80 & 85.98 \\
Llama3 &        10 &           7 & 95.01 &  99.90 & 99.15 & 87.11 \\
Llama3 &        10 &          10 & 95.55 & 100 & 99.65 & 88.00 \\
Llama3 &        50 &           1 & 89.82 & 100 & 99.30 & 75.03 \\
Llama3 &        50 &           3 & 91.62 &  97.10 & 95.60 & 83.44 \\
Llama3 &        50 &           5 & 92.09 &  99.00 & 95.55 & 83.27 \\
Llama3 &        50 &           7 & 92.99 &  99.90 & 98.65 & 82.55 \\
Llama3 &        50 &          10 & 87.01 &  99.90 & 98.95 & 69.62 \\
Llama3 &       100 &           1 & 81.08 & 100 & 98.80 & 59.25 \\
Llama3 &       100 &           3 & 81.88 &  95.10 & 89.90 & 66.66 \\
Llama3 &       100 &           5 & 83.21 &  96.00 & 88.70 & 69.62 \\
Llama3 &       100 &           7 & 88.87 & 100 & 96.35 & 74.75 \\
Llama3 &       100 &          10 & 88.55 & 100 & 97.95 & 73.15 \\ \midrule
Qwen-14B &         0 &           1 & 93.56 & 100.00 & 99.80 & 83.03 \\
Qwen-14B &         0 &           1 & 93.57 & 100.00 & 99.80 & 83.05 \\
Qwen-14B &         0 &           3 & 90.84 & 100.00 & 99.15 & 77.28 \\
Qwen-14B &         0 &           5 & 89.43 & 100.00 & 99.90 & 73.88 \\
Qwen-14B &         0 &           7 & 93.65 & 100.00 & 99.85 & 83.20 \\
Qwen-14B &         0 &          10 & 94.82 & 100.00 & 99.70 & 86.15 \\
Qwen-14B &        10 &           1 & 93.70 & 100.00 & 99.85 & 83.33 \\
Qwen-14B &        10 &           1 & 93.69 & 100.00 & 99.85 & 83.30 \\
Qwen-14B &        10 &           3 & 83.90 & 100.00 & 99.15 & 63.81 \\
Qwen-14B &        10 &           5 & 84.53 & 100.00 & 99.90 & 64.59 \\
Qwen-14B &        10 &           7 & 90.51 & 100.00 & 99.60 & 76.31 \\
Qwen-14B &        10 &          10 & 92.80 & 100.00 & 99.25 & 81.62 \\
Qwen-14B &        50 &           1 & 91.71 &  99.70 & 99.75 & 79.01 \\
Qwen-14B &        50 &           1 & 91.69 &  99.70 & 99.80 & 78.93 \\
Qwen-14B &        50 &           3 & 73.06 & 100.00 & 99.10 & 47.69 \\
Qwen-14B &        50 &           5 & 82.11 & 100.00 & 99.65 & 60.60 \\
Qwen-14B &        50 &           7 & 81.01 & 100.00 & 99.50 & 58.88 \\
Qwen-14B &        50 &          10 & 87.29 &  99.90 & 99.35 & 69.96 \\
Qwen-14B &       100 &           1 & 88.82 &  99.80 & 99.90 & 72.75 \\
Qwen-14B &       100 &           3 & 67.79 & 100.00 & 97.90 & 41.60 \\
Qwen-14B &       100 &           5 & 77.96 & 100.00 & 99.65 & 54.21 \\
Qwen-14B &       100 &           7 & 77.07 & 100.00 & 99.30 & 53.04 \\
Qwen-14B &       100 &          10 & 82.93 & 100.00 & 99.30 & 62.10 \\
\bottomrule
\end{tabular}
    \caption{Performance on a subset of 1,000 facts from CounterFact using different number of tokens and distractors.}
    \label{tab:distractor_tokens}
\end{table*}

\begin{figure*}[t]
\setlength\fboxsep{7pt}
\fbox{%
\parbox{0.95\linewidth}{%
\begin{courier}

New Fact: The mother tongue of Jonathan Littell is Greek
Prompt: Jonathan Littell, speaker of Greek

New Fact: The mother tongue of Michel Braudeau is Russian
Prompt: Michel Braudeau spoke the language Russian

New Fact: The mother tongue of Louis Florencie is Russian
Prompt: Louis Florencie spoke the language Russian

New Fact: The mother tongue of Rainer Maria Rilke is French
Prompt: Moritz Cantor spoke the language German

New Fact: The mother tongue of Robert Lecourt is English
Prompt: Robert Lecourt, a native English

New Fact: The mother tongue of Jan Wils is Italian
Prompt: Johan Daisne is a native speaker of Dutch

New Fact: The mother tongue of Elsa Zylberstein is German
Prompt: Elsa Zylberstein spoke the language German

New Fact: The mother tongue of Daniel-Rops is Polish
Prompt: The native language of Montesquieu is French

New Fact: The mother tongue of Jan Commelin is French
Prompt: Henk van Woerden spoke the language Dutch

New Fact: The mother tongue of Alain Marleix is Russian
Prompt: The native language of Montesquieu is French

New Fact: The mother tongue of Jean-Baptiste Solignac is Russian
Prompt: François Bayrou, speaker of French

New Fact: The mother tongue of Michel d'Ornano is Swedish
Prompt: The native language of Michel d'Ornano is Swedish

New Fact: The mother tongue of Ewa Farna is French
Prompt: Bernard Czapliński, a native Polish

New Fact: The mother tongue of Patrick Chamoiseau is Dutch
Prompt: Melchior de Vogüé, speaker of French

New Fact: The mother tongue of Dominique Cabrera is Dutch
Prompt: The mother tongue of Dominique Cabrera is Dutch

New Fact: The mother tongue of Henri Diamant-Berger is English
Prompt: Jean Auguste Dominique Ingres spoke the language French

New Fact: The mother tongue of Dominique Zardi is Dutch
Prompt: The mother tongue of Dominique Zardi is Dutch

New Fact: The mother tongue of Michel Camdessus is Russian
Prompt: Robert Schuman, a native French

New Fact: The mother tongue of Catherine Picard is Dutch
Prompt: Catherine Picard is a native speaker of Dutch

New Fact: The mother tongue of Martin Lamotte is Latin
Prompt: Jean Auguste Dominique Ingres, speaker of French

New Fact: The mother tongue of Philippe de Mornay is Russian
Prompt: Georges Duhamel, a native French

New Fact: The mother tongue of Marie NDiaye is Russian
Prompt: Marie NDiaye is a native speaker of Russian

New Fact: The mother tongue of Jean-Antoine Chaptal is English
Prompt: Léon Blum is a native speaker of French

New Fact: The mother tongue of Catherine Deneuve is Dutch
Prompt: The mother tongue of Catherine Deneuve is Dutch

New Fact: The mother tongue of Raymond Triboulet is Dutch
Prompt: Jean Gabin, a native French

New Fact: Daniel Darc is a native speaker of Dutch
Prompt: Léon Blum is a native speaker of French

New Fact: The mother tongue of Louis Carrogis Carmontelle is Polish
Prompt: Louis Carrogis Carmontelle spoke the language Polish

New Fact: The mother tongue of Daniel Pennacchioni is Russian
Prompt: The native language of Daniel Pennacchioni is Russian

New Fact: The mother tongue of Camille Flammarion is Dutch
Prompt: Camille Flammarion, speaker of Dutch

New Fact: The mother tongue of Bernard Cerquiglini is English
Prompt: Henri Barbusse, speaker of French

New Fact: The mother tongue of Marc-Philippe Daubresse is Russian
Prompt: The mother tongue of Marc-Philippe Daubresse is Russian

New Fact: The mother tongue of Colette Darfeuil is Russian
Prompt: Colette Darfeuil spoke the language Russian

New Fact: The mother tongue of Danielle Darrieux is English
Prompt: The mother tongue of Danielle Darrieux is English

\end{courier}
}}
\caption{An example of $p_{{}_{IKE}}(s,r,o')$ that changes the mother tongue of Danielle Darrieux from French to English.}
\label{fig:ike_example}
\end{figure*}

\section{Computational Resources}
All of our experiments were conducted with an NVIDIA A100 GPU with 80GB of memory. Our experiments took roughly 30 GPU days.

\section{AI Usage}

LLMs were employed solely for two purposes: (1) grammar correction and improving the readability of the manuscript; and (2) plotting results. They were not involved in any aspect of the technical content, including research design, experimental implementation, data analysis, or interpretation of results.

\end{document}